# Towards Automated Biometric Identification of Sea Turtles (Chelonia mydas)

Irwandi Hipiny[1], Hamimah Ujir[1], Aazani Mujahid[2] & Nurhartini Kamalia Yahya[3]

[1]Faculty of Computer Science and Information Technology, UNIMAS, 94300 Kota Samarahan, Sarawak, Malaysia.
[2]Faculty of Resource Science and Technology, UNIMAS, 94300 Kota Samarahan, Sarawak, Malaysia.
[3]Danau Girang Field Centre, Sabah, Malaysia.

Email: mhihipni@unimas.my

**Abstract.** Passive biometric identification enables wildlife monitoring with minimal disturbance. Using a motion-activated camera placed at an elevated position and facing downwards, we collected images of sea turtle carapace, each belonging to one of sixteen Chelonia mydas juveniles. We then learned co-variant and robust image descriptors from these images, enabling indexing and retrieval. In this work, we presented several classification results of sea turtle carapaces using the learned image descriptors. We found that a template-based descriptor, i.e., Histogram of Oriented Gradients (HOG) performed exceedingly better during classification than keypoint-based descriptors. For our dataset, a high-dimensional descriptor is a must due to the minimal gradient and color information inside the carapace images. Using HOG, we obtained an average classification accuracy of 65%.

## 1 Introduction

Biometric identification of sea turtles within a population is essential for behavioral and ecological study, allowing researchers to estimate vital statistics such as growth rate, survivorship, foraging patterns and population size. Traditional methods of permanent marking and artificial tagging induce stress and possibly harm the animals. Furthermore, tag loss is common due to various factors, namely elapsed time after tagging, study area, target species, size of animal, piercing site and tag's properties (e.g., material, colour and design) [1, 2, 3, 4, 5]. Individual sea turtles can also be recognised via photographic identification of their natural marks, for example, based on coloration patterns around the head area [6], facial profiles [7] and facial scute patterns [8]. Still, the mark-recapture process puts a considerable amount of stress to the animal.

We propose a passive biometric identification system of sea turtles based on robust and co-variant image descriptor matching, see Figure 1. A distant camera remotely captures aerial images of sea turtles at their nesting site. These images are each learned as a robust and co-variant image descriptor, enabling indexing and retrieval. The setup is non-invasive i.e., using a remote camera and no flash photography. Using this setup, we collected images of sixteen juveniles at a private breeding farm in Lundu, Sarawak. The images were taken at night (since female nest at night) inside a perimeter to imitate an actual nesting site. The image descriptor is learned from the most visible part of the sea turtle, as seen aerially, i.e., its carapace. A Chelonia mydas' carapace, see Figure 2, contains a distinctive scute pattern that can be used to identify individuals [9].

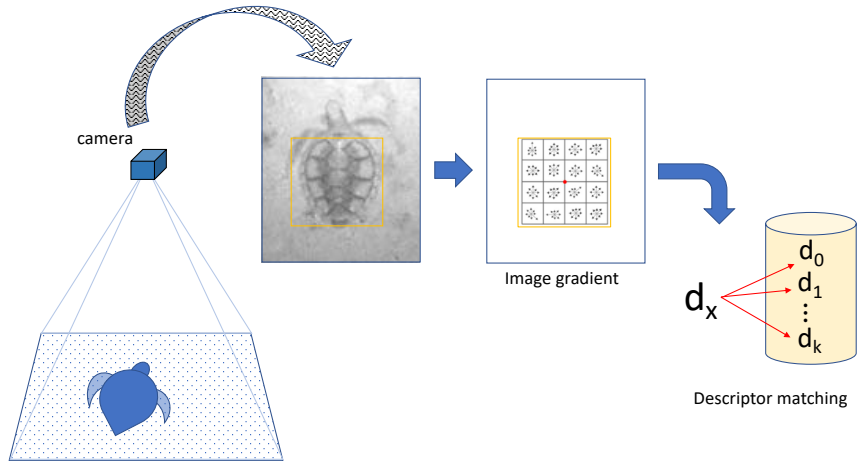

**Figure 1** Our proposed framework. Matching is a minimising function, $\omega(d_x, d_y)$.

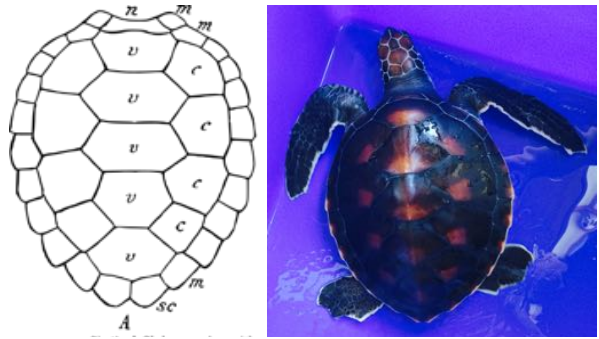

**Figure 2** Left to right. An outline drawing of a Chelonia mydas's carapace, sourced from [10], and an actual image of a juvenile Chelonia mydas kept in a private breeding farm in Lundu, Sarawak.

## 2  Related Work

Recent works on biometric identification of animals, not limited to sea turtles, are motivated by the use of Computer Vision's pattern matching algorithms. An automated approach is a natural progression from manual inspection by human experts. Burghardt et al. [11] used an extended version of Belongie et al.'s Shape Contexts [12] to encode unique phase singularities of spot patterns on individual African penguins. In a more recent work [13], they proposed the detection of shape curls to represent individual Turing-patterned animals. Dababera and Rodrigo [14] implemented an eigenface-based identification mechanism to recognise individual African elephants using their frontal-view face images. Loos and Pfiter [15] combined global and local facial features for visual identification of primates. Taha et al. [16] learned SIFT [17] features from individual horse's muzzle images, later using RANSAC to remove outliers during matching. Also using muzzle images, Monteiro [18] combined graph matching and local invariant features to recognise individual cattle. Li et al. [19] learned Zernike moments from tailhead images of Holstein dairy cows to recognise individuals. To the best of our knowledge, we are the first to develop a passive biometric identification system for recognising individual sea turtles.

### 2.1  Linear Deformation of Scute Pattern

Matching sea turtle individuals based on images captured by a stationary camera is non-trivial due to the linear deformation of salient image features. In such settings, deformation may consist of scale change and in-plane translation and rotation. Dorai et al. [20] suggested several pre-emptive strategies to limit the impact of image deformation when collecting biometric data. Their approach requires multiple measurements to be taken concurrently, later sorted according to the severity of deformity.

Building on a different strategy, image feature descriptors such as [17, 21, 22] and [23], provide a better way to match scute patterns. The gradient-based feature descriptors are co-variant or at least, robust to various image transformations. When paired together with the bag-of-words framework, the setup enables partial matching of models, i.e., using probability to find the nearest match. Due to the low-light image capturing resulting to almost zero colour information, plus the minimal texture on sea turtle carapaces, we theorise that keypoint-based descriptors such as SIFT [17] will not fare well against our dataset. As for template-based descriptors, such as HOG [23], the higher dimensionality should provide a more robust description of the scute pattern. Nevertheless, template-based descriptors are not co-variant to rotation.

## 2.2 Matching of Scute Patterns

Between unique landings, it is probable for an individual sea turtle's carapace to acquire new permanent markings, for example, predation marks, scarring and barnacles. It may also acquire new ephemeral markings, for example, algae, sand particles and dirt. Our feature descriptor must be robust against such noise when matching scute patterns. Existing keypoint-based and template-based feature descriptors should solve this problem by providing a degree of robustness against noise during matching. However, the degree of robustness depends on the properties of the captured images.

## 3 Our Dataset

Table 1  PANDAN-CHELOMY dataset.

| No | Number of Images | No | Number of Images |
|----|------------------|----|------------------|
| 1  | 3                | 9  | 4                |
| 2  | 3                | 10 | 6                |
| 3  | 6                | 11 | 3                |
| 4  | 3                | 12 | 6                |
| 5  | 7                | 13 | 3                |
| 6  | 11               | 14 | 2                |
| 7  | 4                | 15 | 3                |
| 8  | 3                | 16 | 3                |

We collected between two to eleven RGB images (of 3,264 × 2,448 resolution) each for sixteen Chelonia mydas juveniles, see Table 1. A total of 70 images were taken sans flash (as turtles are very sensitive to light). The complete dataset (CC-BY-4.0), see Figure 3, is available on the corresponding author's website. The juvenile sea turtles were kept in captivity inside a private breeding farm in Lundu, Sarawak. During data capture, each individual was placed inside a perimeter and was allowed to move around freely for 3 to 5 minutes. The setup aims to replicate the environment similar to a sea turtle's nesting site. All sixteen individuals, as part of a larger group, were released back to the sea in December 2015 [24].

The dataset was pre-processed prior to classification. Images were converted to greyscale and manually rotated to position the carapace in an upright pose, see Figure 4. The pose correction is required to enable the matching of template-based image descriptors. Inside each image, we manually set an ROI window to exclude most of the background. The remaining background elements inside the ROI are later removed via a smoothing function. The rotated images and ROI information are both included inside our PANDAN-CHELOMY dataset.

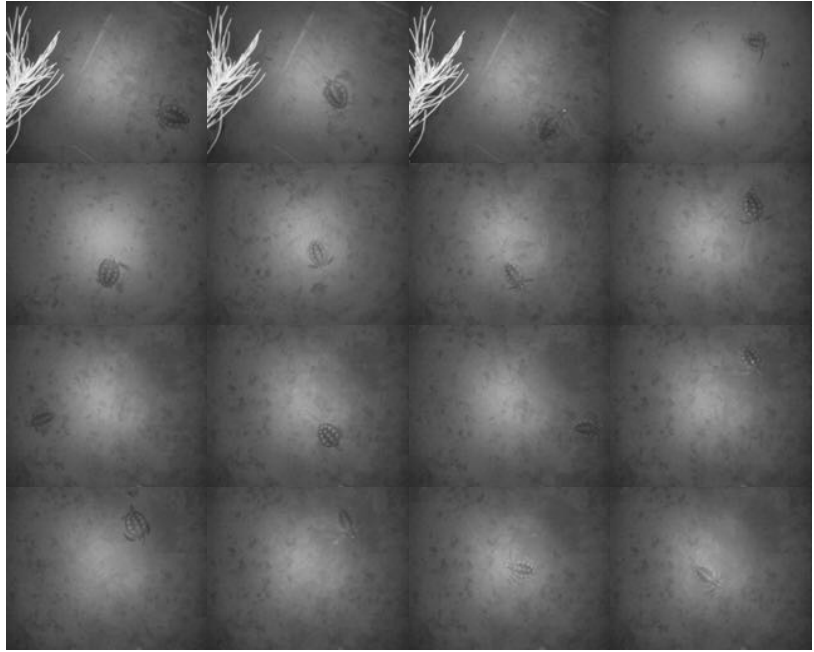

**Figure 3** A From left to right, top to bottom. A sample raw image of each Chelonia mydas juvenile, Turtle 1 to Turtle 16, taken from our dataset PANDAN-CHELOMY.

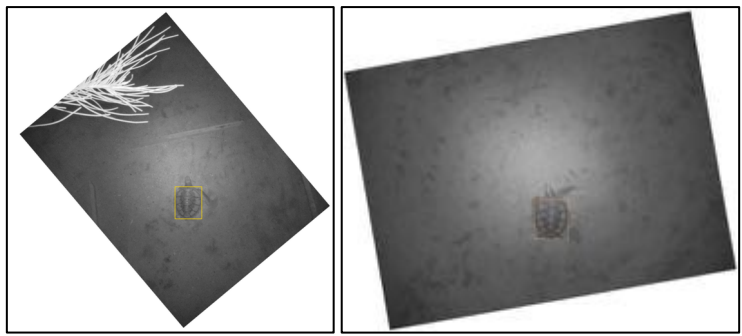

**Figure 4** Rotated images from PANDAN-CHELOMY, each with a visualised ROI.

The dataset was pre-processed prior to classification. Images were converted to greyscale and manually rotated to position the carapace in an upright pose, see Figure 4. The pose correction is required to enable the matching of template-based image descriptors. Inside each image, we manually set an ROI window, see Figure 4, to exclude most of the background. The remaining background elements inside the ROI are later removed via a smoothing function. The rotated images and ROI information are both included inside our PANDAN-CHELOMY dataset.

## 4 Matching of Carapace Images

### 4.1 Classification

The PANDAN-CHELOMY dataset contains 70 images belonging to sixteen juveniles. Prior to matching, the ROIs are smoothed using a $4 \times 4$ Gaussian kernel to suppress the remaining background elements. We found the kernel size to be optimal for removing sand features. A larger kernel erodes more gradient, resulting to a higher loss of discriminative features inside the ROI. A smaller kernel retains more noise, which reduces the classification's accuracy.

Using $k$-fold cross-validation, we obtain matching score for each ROI of the test set against other ROIs of the training set based on the nearest-neighbour distance ratio (NNDR) scheme. The threshold values are varied from 0.0 to 1.0. We calculate the score, $\beta$, as,

$$\beta = \omega_{top}(d_x, d_m) / \omega_{2nd}(d_x, d_n) \tag{1}$$

where $\omega_{top}$ is the distance between the template descriptor, $d_x$, and the best-matching target descriptor, $d_m$, and $\omega_{2nd}$ is the distance between $d_x$ and $d_n$, i.e., the second best-matching target descriptor.

If both $d_x$ and $d_m$ belongs to the same individual, the classification function, $\xi$, returns two possible values,

$$\xi(d_x, d_m) \begin{cases} True\ Positive & \beta \leq threshold \\ False\ Negative & Otherwise \end{cases} \tag{2}$$

Otherwise,

$$\xi(d_x, d_m) \begin{cases} \text{False Positive} & \beta \leq \text{threshold} \\ \text{True Negative} & \text{Otherwise} \end{cases} \quad (3)$$

In the event of the classification function returning multiple top matches, we count the result as a False Negative. Based on the total number of True Positives (TP), False Positives (FP), True Negatives (TN) and False Negatives (FN), obtained from our classification exercise, the True Positive Rate, $TPR = TP/(TP+FN)$ and the False Positive Rate, $FPR = FP/(FP+TN)$, for each threshold value are estimated.

### 4.2 Image Descriptors

We selected Scale-invariant Feature Transform (SIFT) [17], Speeded Up Robust Features (SURF) [21] and Oriented FAST and Rotated BRIEF (ORB) [22], as our keypoint-based descriptors. Histogram of Oriented Gradients (HOG) [23] was chosen as our sole template-based descriptor. All parameters were set to the default values as suggested in the original publications [17] [21] [22] except for HOG.

For HOG, we rescale each ROI to $96 \times 128$ resolution, which translates to a descriptor length of $\left(\frac{96}{8} - 1 \times \frac{128}{8} - 1\right) \times (2 \times 2) \times = 5{,}940$, with $8 \times 8$ pixel cells, blocks of $2 \times 2$ cells, and a 9-bin orientation histogram (0° - 180°). Our ROIs are larger than the $64 \times 128$ resolution used in [23] due to the typical dimension of a sea turtle carapace being almost equal in width and height.

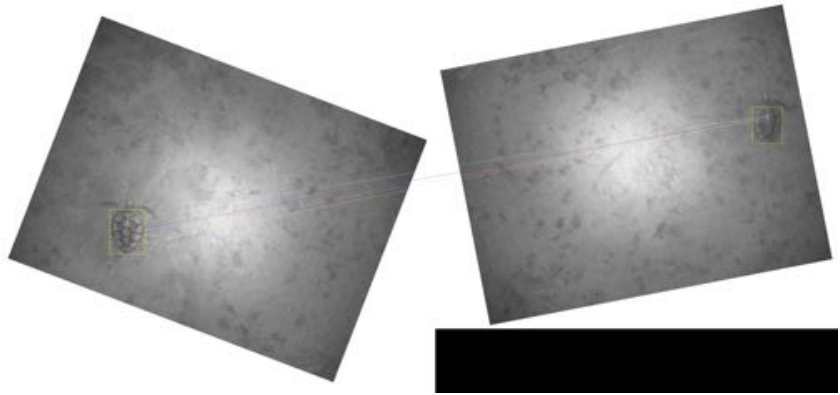

**Figure 5** An example of a SIFT matching result (acceptance threshold of 0.8) between Turtle 8a and Turtle 6d. The number of positive matches is 5. Image brightness was increased to improve visual clarity.

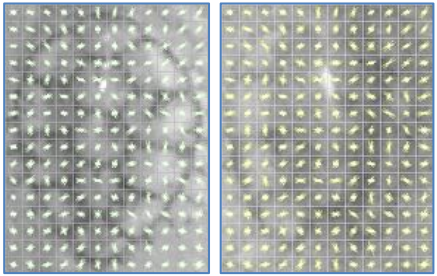

**Figure 6** Visualised HOG descriptor for Turtle 8a (left) and Turtle 6d (right). The L2-norm value is 6.82. Image brightness was increased to improve visual clarity.

Additionally, for SIFT, SURF and ORB, we vary the acceptance threshold during keypoint matching, from 0.2 to 0.8. We plot the classification result for each acceptance threshold separately. See Figure 5 for an example of a SIFT matching result and Figure 6 for an example of visualised HOG descriptors for two paired carapace images.

## 5    Results and Discussion

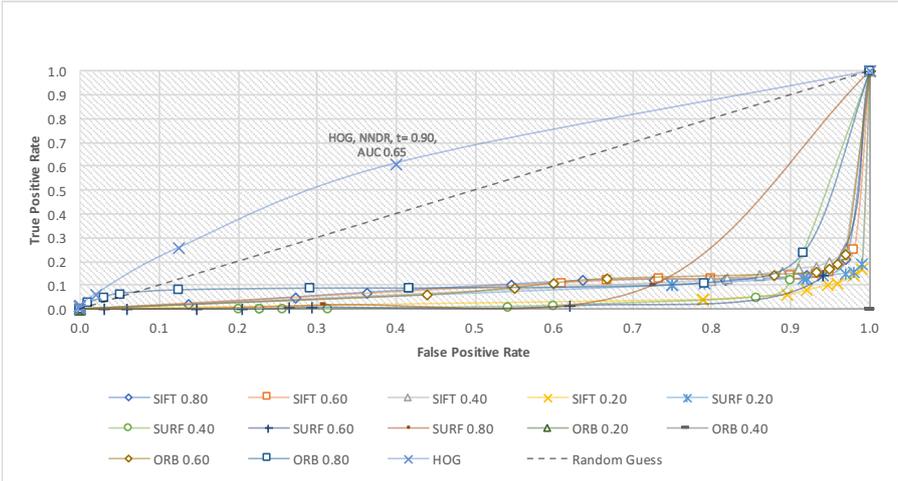

**Figure 7** ROC curve plots obtained from all classification results.

| | Turtle 1 | Turtle 2 | Turtle 3 | Turtle 4 | Turtle 5 | Turtle 6 | Turtle 7 | Turtle 8 | Turtle 9 | Turtle 10 | Turtle 11 | Turtle 12 | Turtle 13 | Turtle 14 | Turtle 15 | Turtle 16 | MANY |
|---|---|---|---|---|---|---|---|---|---|---|---|---|---|---|---|---|---|
| Turtle 1 | 1.00 | 0.00 | 0.00 | 0.00 | 0.00 | 0.00 | 0.00 | 0.00 | 0.00 | 0.00 | 0.00 | 0.00 | 0.00 | 0.00 | 0.00 | 0.00 | 0.00 |
| Turtle 2 | 0.00 | 1.00 | 0.00 | 0.00 | 0.00 | 0.00 | 0.00 | 0.00 | 0.00 | 0.00 | 0.00 | 0.00 | 0.00 | 0.00 | 0.00 | 0.00 | 0.00 |
| Turtle 3 | 0.00 | 0.00 | 0.33 | 0.00 | 0.00 | 0.00 | 0.00 | 0.00 | 0.00 | 0.00 | 0.17 | 0.00 | 0.00 | 0.00 | 0.00 | 0.00 | 0.50 |
| Turtle 4 | 0.00 | 0.00 | 0.00 | 0.67 | 0.00 | 0.00 | 0.00 | 0.00 | 0.00 | 0.00 | 0.00 | 0.00 | 0.33 | 0.00 | 0.00 | 0.00 | 0.00 |
| Turtle 5 | 0.00 | 0.00 | 0.00 | 0.00 | 0.14 | 0.00 | 0.00 | 0.00 | 0.00 | 0.00 | 0.00 | 0.00 | 0.29 | 0.00 | 0.00 | 0.00 | 0.57 |
| Turtle 6 | 0.00 | 0.00 | 0.00 | 0.00 | 0.09 | 0.09 | 0.00 | 0.00 | 0.00 | 0.00 | 0.00 | 0.00 | 0.00 | 0.00 | 0.09 | 0.00 | 0.73 |
| Turtle 7 | 0.00 | 0.00 | 0.00 | 0.00 | 0.25 | 0.00 | 0.50 | 0.00 | 0.00 | 0.00 | 0.00 | 0.00 | 0.00 | 0.00 | 0.00 | 0.00 | 0.25 |
| Turtle 8 | 0.00 | 0.00 | 0.00 | 0.00 | 0.00 | 0.00 | 0.00 | 1.00 | 0.00 | 0.00 | 0.00 | 0.00 | 0.00 | 0.00 | 0.00 | 0.00 | 0.00 |
| Turtle 9 | 0.00 | 0.00 | 0.00 | 0.00 | 0.00 | 0.00 | 0.00 | 0.00 | 0.75 | 0.00 | 0.00 | 0.00 | 0.00 | 0.00 | 0.00 | 0.00 | 0.25 |
| Turtle 10 | 0.00 | 0.00 | 0.00 | 0.00 | 0.00 | 0.00 | 0.00 | 0.00 | 0.00 | 0.50 | 0.00 | 0.00 | 0.00 | 0.00 | 0.00 | 0.00 | 0.50 |
| Turtle 11 | 0.00 | 0.00 | 0.00 | 0.00 | 0.00 | 0.00 | 0.00 | 0.00 | 0.00 | 0.00 | 1.00 | 0.00 | 0.00 | 0.00 | 0.00 | 0.00 | 0.00 |
| Turtle 12 | 0.00 | 0.00 | 0.00 | 0.00 | 0.00 | 0.00 | 0.00 | 0.00 | 0.00 | 0.00 | 0.00 | 0.50 | 0.00 | 0.00 | 0.00 | 0.00 | 0.50 |
| Turtle 13 | 0.00 | 0.00 | 0.00 | 0.00 | 0.00 | 0.00 | 0.00 | 0.00 | 0.00 | 0.00 | 0.00 | 0.00 | 1.00 | 0.00 | 0.00 | 0.00 | 0.00 |
| Turtle 14 | 0.00 | 0.00 | 0.50 | 0.00 | 0.00 | 0.00 | 0.00 | 0.00 | 0.00 | 0.00 | 0.00 | 0.00 | 0.00 | 0.00 | 0.00 | 0.00 | 0.50 |
| Turtle 15 | 0.00 | 0.00 | 0.00 | 0.00 | 0.00 | 0.00 | 0.00 | 0.00 | 0.00 | 0.00 | 0.00 | 0.00 | 0.00 | 0.00 | 1.00 | 0.00 | 0.00 |
| Turtle 16 | 0.00 | 0.00 | 0.00 | 0.00 | 0.00 | 0.00 | 0.00 | 0.00 | 0.00 | 0.00 | 0.00 | 0.00 | 0.00 | 0.00 | 0.00 | 1.00 | 0.00 |

Actual classes (rows); Predicted classes (columns)

**Figure 8** Confusion matrix obtained from the classification results using HOG [23], with threshold value for NNDR scheme set to 0.9.

Our classification of 70 sea turtle carapaces, using $k$-fold cross-validation, produced the ROC curve plots shown in Figure 7. As predicted, all keypoint-based descriptors, with acceptance threshold values ranging from 0.2 to 0.8, had failed spectacularly with worse performance than random guessing. Only HOG had performed better (than random guessing) with an average accuracy of 65%. Classification accuracy of this dataset via random guessing is 6.25%. Evidently, the nature of our ROI images, i.e., minimal colour information and lack of textures, contributes to the failure of SIFT, SURF and ORB. Our implementation of HOG produces a descriptor length of 5,940, which is a far greater number than SIFT's 128, making it more discriminative and robust.

The optimal threshold value for HOG is 0.9. This reveals that with HOG, even though we managed to obtain an average classification accuracy of 65%, the distance between the top match and the second-best match is nominal. The confusion matrix using the optimal configuration is shown in Figure 3(b). We have 16 actual classes and $16+1$ predicted classes. The additional predicted class, i.e., class MANY, represents cases where our classification function returns multiple top matches. We consider such cases as a False Negative to penalise the configuration.

## 6     Conclusion

Based on our results, we conclude that the recognition of Chelonia mydas individuals using aerial images of their carapace is possible. By learning these ROI images as HOG descriptors, we managed to obtain an average classification accuracy of 65%, with certain individuals managing 75% or higher. By dealing with cases where multiple top matches are returned from a single classification instance, we should be able to improve the average classification accuracy further. Solutions such as cumulative voting scheme [25] [26] and modular classification [27] shall be explored in the future. Another potential solution is to use 3D features captured using an RGB-D sensor to represent the scutes, such as surface normal [28].

Constrained by our grant's limitation, we admittedly ignored the effect of scute deformations over time. The dataset contains images captured during a single landing. In future, we plan to collect images of multiple landings at an actual nesting site over a longer period of time.